\pdfoutput=1
\documentclass{article}
\usepackage{CJKutf8}
\usepackage{INTERSPEECH2022}
\usepackage{amsmath,graphicx,multirow,url,adjustbox}
\usepackage{spconf}
\usepackage{cite}
\usepackage{color,soul}
\usepackage{hyperref}
\usepackage{mathtools}
\usepackage{tabularx}
\usepackage{booktabs}
\usepackage{stfloats}
\usepackage[pdf]{graphviz}
\usepackage{pgfplots}
\pgfplotsset{width=\linewidth*1.04,compat=1.17}
\usepackage{subcaption}
\usepackage{listings}
\usepackage{url}
\definecolor{our_green}{HTML}{76b900}
\definecolor{our_blue}{HTML}{6d9eeb}

\usepackage{xpatch}
\makeatletter
\newcommand*{\addFileDependency}[1]{
  \typeout{(#1)}
  \@addtofilelist{#1}
  \IfFileExists{#1}{}{\typeout{No file #1.}}
}
\makeatother
\xpretocmd{\digraph}{\addFileDependency{#2.dot}}{}{}

\newcolumntype{Y}{>{\centering\arraybackslash}X}

\title{Transducers with Pronunciation-aware Embeddings for Automatic Speech Recognition}

\name{\begin{tabular}{c} Hainan Xu, Zhehuai Chen, Fei Jia, Boris Ginsburg
\end{tabular}
}
\address{
  NVIDIA, USA \\
\small{\url{hainanx@nvidia.com}} }

\newcommand\blfootnote[1]{%
  \begingroup
  \renewcommand\thefootnote{}\footnote{#1}%
  \addtocounter{footnote}{-1}%
  \endgroup
}

\begin{document}

\maketitle

\begin{abstract}
This paper proposes \emph{Transducers with Pronunciation-aware Embeddings} (PET). Unlike conventional Transducers where the decoder embeddings for different tokens are trained independently, the PET model's decoder embedding incorporates shared components for text tokens with the same or similar pronunciations. With experiments conducted in multiple datasets in Mandarin Chinese and Korean, we show that PET models consistently improve speech recognition accuracy compared to conventional Transducers. Our investigation also uncovers a phenomenon that we call \emph{error chain reactions}. Instead of recognition errors being evenly spread throughout an utterance, they tend to group together, with subsequent errors often following earlier ones. Our analysis shows that PET models effectively mitigate this issue by substantially reducing the likelihood of the model generating additional errors following a prior one. Our implementation will be open-sourced with the NeMo toolkit.

\end{abstract}


\section{Introduction}
Over the last decades, research in speech recognition has experienced a shift in the prominent model architectures for the task.
Previously, the most successful speech recognition models used a modular/hybrid approach \cite{povey2011kaldi,young2002htk},
where an acoustic model, a language model, and a pronunciation model are independently built from data, and combined together to build a speech recognition pipeline. Nowadays, end-to-end speech recognition models \cite{graves2006connectionist,graves2012sequence,chorowski2015attention,chan2016listen} have gradually surpassed the performance of hybrid models. End-to-end ASR uses a single model that takes in the acoustic input, and directly outputs the recognized results. 
Inside end-to-end models, there are no separate acoustic/language/pronunciation modules. Instead, all information regarding different levels of representation of speech and languages is internally learned. 
Despite their simplicity, end-to-end models have shown superior performance than hybrid models, and are widely supported in the open-source community\cite{kuchaiev2019nemo,ren2019fastspeech,wang2019espresso,ravanelli2021speechbrain}. \blfootnote{Paper accepted at ICASSP 2024 conference.}

Acknowledging the overall success of end-to-end models, there are still a number of aspects where hybrid models are more attractive. For example, because all the information is implicitly learned without linguistic knowledge, end-to-end models are usually much more data-hungry than hybrid models in order to learn the intricacy of languages. One such area is regarding the pronunciation information of text tokens. 
Hybrid models typically have access to a pronunciation dictionary, which makes tokens with the same or similar pronunciations share certain parameters;
On the end hand, end-to-end models typically treat different word tokens as completely independent tokens, even if they share the same or similar pronunciations, this could lead to sub-optimal results\cite{shen2023pronunciation}.

There has been ongoing investigations on how to leverage pronunciation information to improve end-to-end models. For example, \cite{9688053} and \cite{wang2021cascade} both proposed a 2-stage ASR model, by first recognizing phonetic symbols which is then converted into Chinese characters. 
\cite{xu2019improving} proposed a subword tokenization scheme using the pronunciation information of the words to guide the subword splitting process. This method was shown to be able to identify phonetically meaningful subword tokens and achieved superior results than conventional BPE tokenization.
\cite{pandey2023procter} proposed a pronunciation-aware mechanism which is included in the contextual adapter framework \cite{chang2021context} for Transducer models. 
Both \cite{chen2019joint} and \cite{bruguier2019phoebe} proposed a pronunciation-aware contextualization method on top of the Contextual-Listen-Attend-and-Spell \cite{pundak2018deep} framework.
\cite{shen2023pronunciation} proposed to represent Mandarin characters with their pronunciation and a unique character ID, in order to promote parameter sharing among characters with the same or similar pronunciations, and achieved improved ASR results.

In this paper, we propose \emph{Transducers with Pronunciation-aware Embeddings} (PET), an extension to Transducer \cite{graves2012sequence} that can encode expert knowledge of a pronunciation dictionary in the model's embedding parameters. 
To the best of our knowledge, our work is the first to incorporate pronunciation information of text tokens into the \emph{embedding design} of Transducer models.
The contributions of this paper are,
\begin{enumerate}
    \item a novel embedding generation scheme for Transducer decoders, where embeddings for text tokens can have shared components based on the similarity of their pronunciations. 
    \item the discovery of \emph{error chain reactions} in Transducers, where errors tend to group instead of being evenly distributed, and one error is likely to cause other subsequent errors. At a time when overall ASR error rates are gradually getting saturated, this discovery allows researchers to develop specific methods (of which this paper is one) targeting error chains to further improve ASR models' performance.

    \item PET models improve ASR consistently for Mandarin Chinese and Korean. Furthermore, PET models outperform conventional models primarily by mitigating error chain reactions, reducing the likelihood of more errors after a prior error.    
\end{enumerate}

 We will open-source our model with the NeMo \cite{kuchaiev2019nemo} toolkit.

\begin{figure}
    \centering
    \includegraphics[scale=0.3]{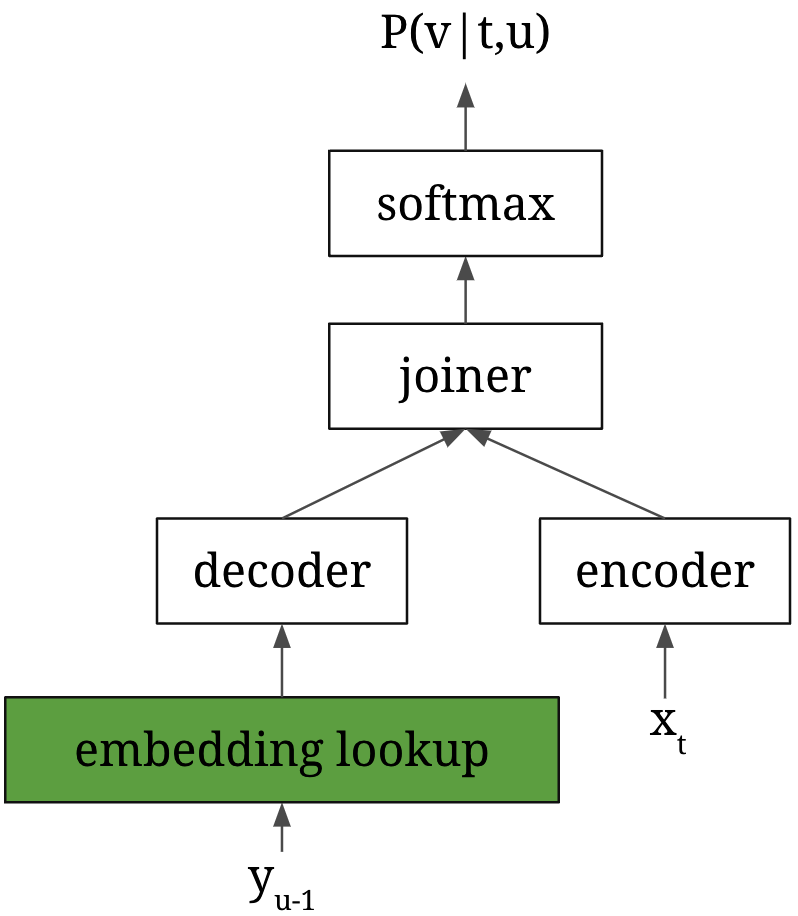}
    \caption{Transducer model architecture.}
    \label{rnnt}
\end{figure}

\section{Background and Motivation for PET}

The Transducer model is one of the popular types of end-to-end ASR models \cite{graves2012sequence}. The model improves upon CTC \cite{graves2006connectionist} models by relaxing the conditional independence assumption, and has a separate encoder and decoder to process the audio and previously emitted text. The learned high-level information from the encoder and decoder is fed into a joiner (also called a joint network) to generate token distributions based on both the acoustics and emission history.  The architecture of Transducers is shown in Figure \ref{rnnt}, where $x$ represents the acoustic input indexed by $t$, and $y$ represents the text token indexed by $u$. $P(v | t, u)$ represents the model's prediction of the output distribution when accessing $t$'s frame of acoustic input, given the previous $u - 1$ emissions.

\begin{figure}
    \centering
    \includegraphics[scale=0.34]{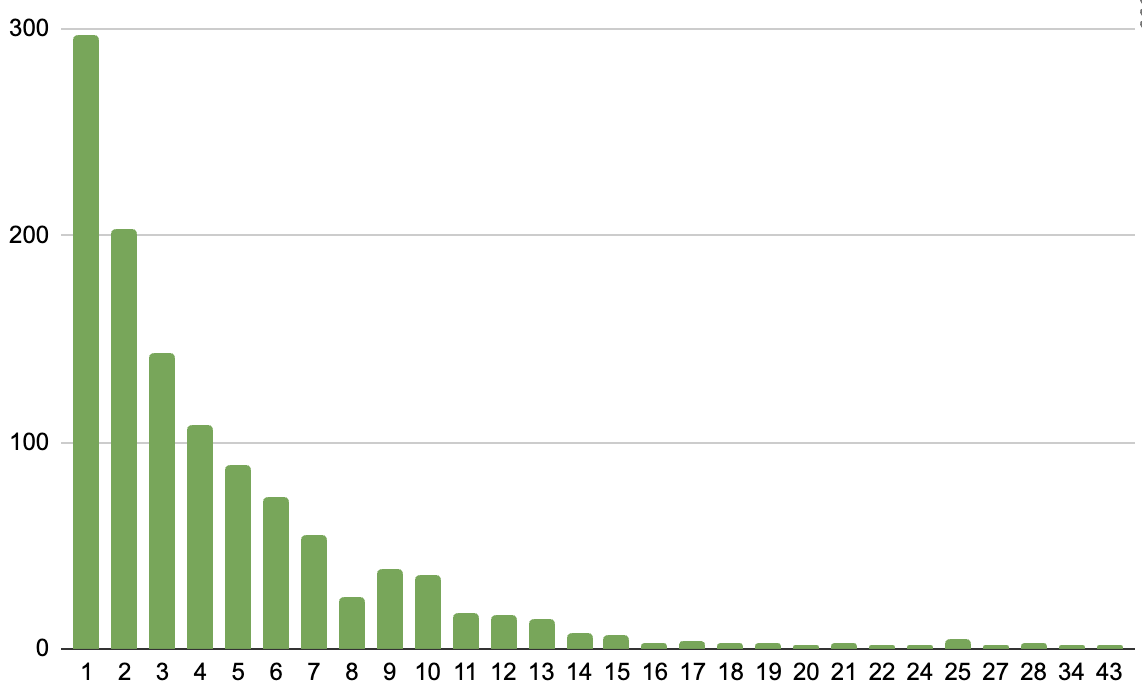}
    \caption{A Chart showing homophone distributions in Mandarin Chinese, where a bar $(x, y)$ means the number of different pronunciations shared by exactly $x$ characters is $y$. E.g. there are just below 300 characters that have unique pronunciations; slightly above 200 pronunciations are shared by two Chinese characters. On the right-hand side of the chart, we see certain pronunciations can be shared among as many as 43 characters.}
    \label{frequency}
\end{figure}

Unlike the acoustic input which resides in a continuous vector space, text tokens are discrete symbols and thus need to be projected onto a continuous vector space before decoder computation. In Transducers, an embedding layer of dimension $[V, d]$ is required for the decoder, where $V$ is the vocabulary size and $d$ is the embedding dimension.
With conventional Transducers, embeddings with respect to different tokens are independently trained, despite the fact that different tokens can be linguistically related. One such example is Mandarin Chinese, where there exists a large number of \emph{homophones}, i.e. different characters that share the same pronunciation. For example in AISHELL-2 dataset\cite{du2018aishell}, its 5000 characters have in total of 1149 different pronunciations\footnote{We use \emph{pinyin} package \url{https://pypi.org/project/pinyin/} to generate the pronunciations for Mandarin Chinese in this paper.}. Fig.\ref{frequency} shows more detailed homophone patterns in Mandarin Chinese, 
where a green bar $(x, y)$ means there are exactly $y$ pronunciations that are shared by exactly $x$ Mandarin Chinese characters.
We see just under 300 of all 5000 characters have unique pronunciations and all the rest are homophones.
We believe it's sub-optimal to treat homophones as independent tokens and propose to incorporate shared components in embeddings for tokens with the same or similar pronunciations.

\section{Pronunciation-aware Embeddings}

Different from conventional Transducers that use simple word embedding matrices, embeddings for PET models are a combination of multi-faceted representations of the text tokens. To do this, PET models can have multiple embedding tables, some of which correspond to pronunciation symbols. 
To compute the \emph{final embedding} (the actual embedding passed into decoders) for a word $v$, we first choose the features we want to use for generating the embedding. We propose features shown in Table \ref{shorthand} to represent different aspects of the word.
Say $F$ is the set of features we choose, then the final embedding for a word $v$ is computed as the summation \footnote{Readers might feel that concatenating those embeddings would be a more logical design. We point out it is just a special case of adding, where the individual components lie in orthogonal subspaces. Our preliminary experiments show that it doesn't work as well as adding.} of embeddings of different features of that word,
\begin{equation}
    \bold{E}_\text{FINAL, F}(v) = \sum_{f \in F} \bold{E}_\text{f}(f(v))
    \label{pet}
\end{equation}
Here are some examples to illustrate how this works. If we choose features \{P, W\}, then the final embedding for characters \text{\begin{CJK*}{UTF8}{gbsn}他\end{CJK*}} and \text{\begin{CJK*}{UTF8}{gbsn}她\end{CJK*}}, both of which are pronounced ``ta'', are computed as,
\begin{equation}
 \bold{E}_\text{FINAL, PW}(\text{\begin{CJK*}{UTF8}{gbsn}他\end{CJK*}}) = \bold{E}_\text{p}(\text{ta}) + \bold{E}_\text{w}(\text{\begin{CJK*}{UTF8}{gbsn}他\end{CJK*}}) 
\end{equation}
\begin{equation}
\bold{E}_\text{FINAL, PW}(\text{\begin{CJK*}{UTF8}{gbsn}她\end{CJK*}}) = \bold{E}_\text{p}(\text{ta}) + \bold{E}_\text{w}(\text{\begin{CJK*}{UTF8}{gbsn}她\end{CJK*}}) 
\end{equation}
In this example, $\bold{E}_p$ and $\bold{E}_w$ are two independent embedding tables corresponding to different representations of the text tokens and will be trained jointly in Transducer training.
The choice of $F$ can be arbitrary with defined features. It is also OK to drop the ``W'' feature in the embedding generation so different tokens can have their embeddings completely tied.
For example, when using features ``CV'',  the embeddings for \text{\begin{CJK*}{UTF8}{gbsn}他\end{CJK*}} and \text{\begin{CJK*}{UTF8}{gbsn}她\end{CJK*}} would be identical,
\begin{equation}
 \bold{E}_\text{FINAL, CV}(\text{\begin{CJK*}{UTF8}{gbsn}他\end{CJK*}}) = \bold{E}_\text{FINAL, CV}(\text{\begin{CJK*}{UTF8}{gbsn}她\end{CJK*}}) = \bold{E}_\text{c}(\text{t}) + \bold{E}_\text{v}(\text{a})
\end{equation}

\begin{table}
    \caption{Short-hands used for denoting embedding features.}
    \centering
    \begin{tabular}{c p{2.3in} c}
    \toprule
    short-hand & description &  example \\
    \midrule
     W    & word identity & \begin{CJK*}{UTF8}{gbsn}他\end{CJK*} \\
     P    & Romanized pronunciation & ta \\
     T   & tone (only for tonal languages) & 1st tone \\
     C   & beginning consonant(s) of P (could be empty if pronunciation begins with vowels) & t \\
     V  & suffix of P excluding the part from C & a \\
     \bottomrule
    \end{tabular}

    \label{shorthand}
\end{table}

\section{Experiments}
We evaluate PET models in Mandarin Chinese and Korean using NeMo \cite{kuchaiev2019nemo}. Our baseline uses the Fast Conformer model \cite{rekesh2023fast}\footnote{  \url{examples/asr/conf/fastconformer/fast-conformer_transducer_bpe.yaml} in the NeMo repository.}.
All models use character-based representations and are trained for sufficient steps till the validation error rate starts to increase. After training, we run checkpoint averaging on the 5 checkpoints with the best validation performance to generate the model for evaluation.

\subsection{Mandarin Chinese experiments}

We train Mandarin Chinese models with the AISHELL-2 \cite{du2018aishell} dataset, which contains 1000 hours of speech. We use the default fast-conformer-transducer configurations, with  120M parameters. 
We report our models'  \emph{character error rates} (CER) performance on AISHELL-2's iOS-test set and THCHS's testset \cite{wang2015thchs} in Table \ref{chinese_wer}.
Comparing baseline (W) with the PW model, we see that adding pronunciation information on top of word identity in the embeddings does not help improve ASR performance; however, including \emph{only pronunciation information} in the decoder embeddings can reduce the CER of PET models. Also, it seems a more granular representation of the pronunciation can yield more gains, e.g. using CV or V would outperform P. This effect is more pronounced for the out-of-domain THCHS dataset. Our best model uses the suffix of the pronunciation after the initial consonants and achieved a relative CER reduction of 2.7\%  for the AISHELL iOS-test dataset, and an absolute reduction of 1.01\% (or 7.1\% relative) on the THCHS test.



\begin{table}
    \caption{CER performance of different Transducers on iOS-test, and THCHS test sets in Mandarin Chinese. All models are trained on AISHELL-2. Check Table \ref{shorthand} for the meaning of representations of configs.
    We see PET models can improve ASR accuracy, especially for the more difficult THCHS dataset, and a more granular representation of the pronunciation is used.}
    \centering
    \begin{tabular}{c c c}
    \toprule
    decoder-emb & iOS-test & THCHS-test  \\
    \midrule
    W (baseline) & 5.50 & 14.16    \\
    \midrule
    PW  & 5.50 & 14.69 \\
    P   & 5.54 & 13.66 \\
    PT  & 5.45 & 13.16 \\
    CV  & 5.55 & 13.23 \\
    V  & \textbf{5.34} & \textbf{13.15} \\
    \bottomrule
    \end{tabular}
    \label{chinese_wer}
\end{table}

\subsection{Korean experiments}
\begin{table}
    \caption{CER of different Transducers on Zeroth-Korean test set. Similar to Table \ref{chinese_wer}, using granular representation of pronunciation for decoder embedding can improve ASR performance.}
    \centering
    \begin{tabular}{c c c c}
    \toprule
      decoder-emb     & zeroth test  \\
    \midrule
        W  & 1.51  \\
        \midrule
        P  & 1.55   \\
        CV  & 1.43 &  \\
        V & \textbf{1.36}  \\
        
    \bottomrule
    \end{tabular}

    \label{korean_wer}
\end{table}

Our Korean models are trained on the Zeroth-Korean \footnote{https://www.openslr.org/40/} dataset, with around 50 hours of speech. Due to the small data size, we decreased the hidden dimension of the encoder to 400 from the default 640, and the model has around 50 million parameters in total. \emph{ko-pron} package\footnote{\url{https://pypi.org/project/ko-pron/}} is used to generate pronunciation for Korean characters. Note that Korean isn't a tonal language, so there is no tone feature.
Table \ref{korean_wer} shows CER on the Zeroth-Korean test set.
We see very similar trends where better results can be achieved when the decoder only uses granular representations of pronunciations (CV or V).


        



\section{Discussions}

\subsection{Pronunciation Information in Joiner Embeddings}
The Transducer joiner has a [V+1, d] matrix which projects hidden representations onto logits of dimension V + 1 (+1 for blank). This matrix could be viewed as the ``joiner embedding'', and we investigate including pronunciation information here as well. Table \ref{chinese_wer_joiner} shows the results in Mandarin Chinese. We see that, generally, including pronunciation only in joiner embedding does not help ASR performance; meanwhile, including pronunciation for both decoder and joiner can bring gains to ASR performance compared to baselines, but doesn't outperform our best number from Table \ref{chinese_wer} which only include pronunciation in decoder embeddings.

As for Korean, the results are in \ref{korean_wer_joiner}. Here we see that different from Mandarin Chinese models, by combining the use of pronunciation in both decoder and joiner, our Korean ASR performance can be further improved. Our best number for the Zeroth-Korean dataset is 1.22\% using V-VW config, which is to our best knowledge, the best reported CER on the dataset. We hypothesize the different patterns for the two languages are due to the data size difference. Since our Korean data is much smaller, including pronunciation for both decoder and joiner embeddings can make the model learn more robust representations. We will continue investigating this as future work.

\begin{table}
    \caption{CER of different Transducers on iOS-test, and THCHS test sets in Mandarin Chinese when model uses pronunciation for joiner embeddings. We see that including pronunciation in joiner embeddings generally doesn't help improve ASR performance.}
    \centering
    \begin{tabular}{c c c c c}
    \toprule
    decoder-emb & joiner-emb & iOS-test & THCHS-test  \\
    \midrule
    W & W  (baseline) & 5.50 & 14.16   \\
    \midrule
    W & PW & 5.62 & 14.55 \\
   W & PTW & 5.59 & 14.37 \\
   W& CVW & 5.56 & 14.47 \\
    \midrule
    P & PW & 5.37 & 13.33 \\
    CV&CVTW & 5.49 & 13.39\\ 
    \bottomrule
    \end{tabular}
    \label{chinese_wer_joiner}
\end{table}

\begin{table}
    \caption{CER of Zeroth-Korean test set with Transducers where the joiner embedding includes pronunciation information.}
    \centering
    \begin{tabular}{c c c }
    \toprule
      decoder-emb & joiner-emb & zeroth  \\
    \midrule
        W & W & 1.51   \\
    \midrule
        W & PW & 1.42 \\
       W & CVW & 1.39  \\
    \midrule
       CV & CVW & \textbf{1.22}   \\
        V & VW & 1.27   \\
       P& PW & 1.33   \\
        
    \bottomrule
    \end{tabular}

    \label{korean_wer_joiner}
\end{table}

        



\subsection{Error Chain Reactions}
We uncover an \emph{error chain reactions} phenomenon when we analyze the error patterns of our models. 
We compute the error rates of each character in the target sentence, conditioned on whether its previous token is recognized correctly. For the special case where the character is at the beginning of the utterance, we assume the ``previous token'' is correct. Table \ref{conditional_cer} shows our results, where $P(E | E)$ and $P(E | c)$ mean the probability of error when the previous character is an error (E) or correct (C), respectively.

\begin{table}[t]
    \caption{Probability of error conditioned on if the previous word is correct. $P(E | E)$ and $P(E | C )$ represent the probability of error when the previous word is an error (E) or correct (C), respectively.}
    \centering
    \begin{tabular}{c c c c c c}
    \toprule
    & & \multicolumn{2}{c}{AISHELL} & \multicolumn{2}{c}{THCHS} \\
    decoder-emb & joiner-emb  & $P(E | E)$ & $P(E | C )$ & $P(E | E)$ & $P(E | C )$  \\
    \midrule
    W & W & 28.7 & 4.2 &  38.1 & 10.2 \\
    \midrule
    W & PW & 28.6 & 4.3 & 38.5 & 10.4 \\
    W & PTW & 28.4 & 4.3 & 37.6 & 10.4  \\
    W & CVW & 29.4 & 4.2 & 38.1 & 10.4  \\
    PW & W & 27.5 & 4.3 & 38.9 & 10.5  \\
    \midrule
    P & W & 24.6 & 4.5 & 33.0 & 10.5 \\
    V & W & \textbf{23.4} & 4.3 & 32.2 & 10.2 \\
    CV & W & 24.3 & 4.5 & 32.0 & 10.3 \\
    P & PW & 23.8 & 4.4 & 32.6 & 10.3 \\
    CV & CVTW & 24.8 & 4.4 & \textbf{31.9} & 10.5 \\
    \bottomrule
    \end{tabular}

    \label{conditional_cer}
\end{table}

From Table \ref{conditional_cer}, we see clear discrepancies between $P(E | E)$ and $P(E | C)$, in all models and all datasets -- in all cases, $P(E | E)$ are always several times larger than $P(E | C)$. This means, if an error occurs with Transducer models, it will greatly increase the chance of the next token being an error too, causing a possible \emph{chain reaction} of errors to occur. We hypothesize that this has to do with the auto-regressive nature of Transducer models, where the emission of a word depends on the emission histories; meanwhile, Transducer model training can suffer from exposure bias \cite{schmidt2019generalization}, where during the model training, it is always using the correct tokens as history; while during inference, the model is fed previously emitted tokens, which might contain errors, in which case, the model is more likely to produce more errors since this is a situation not seen during training.

Conceptually, error chain reaction is an easy-to-understand phenomenon for all auto-regressive models, nevertheless, to the best of our knowledge, this paper is the first to quantitatively study its effect. At a time when most research work targets improving the overall performance of ASR models, and performance numbers are gradually getting saturated, we believe mitigating the effect of error chain reactions is an area that deserves more attention among researchers. 

\subsection{PET Models Suppress Error Chain Reactions}

PET models are more robust to error chain reactions, because when ASR models make errors, the result is usually acoustically similar to the correct word -- for example, the model predicting \begin{CJK*}{UTF8}{gbsn}他\end{CJK*} instead of \begin{CJK*}{UTF8}{gbsn}她\end{CJK*}, which are homophones. As a result, when a PET model produces an error, the error may have all or part of the pronunciation matching that of the correct token, so the decoder embedding might still be correct.
For example, if only V is used in the decoder embedding, and the model produces an error, as long as the error character shares the same pronunciation symbol in V, the decoder will take the ``correct'' embedding and it will not likely cause another error.
This is corroborated in Table \ref{conditional_cer}.
Note, that the experiments are divided into two groups depending on whether word identity information is used in decoder embeddings.
We see from the Table that for both datasets, $P(E | C)$ are relatively similar across all models; however, there is a significant reduction in $P(E | E)$ when W is not used in decoder embedding for both datasets, as much as 5.3\% absolute for AISHELL (V-W) and 6.2\% absolute for THCHS (CV-CVTW). This means PET models can mitigate the issue of error chain reaction. 

We also report the \emph{average length of error clusters} in the data. By an \emph{error cluster}, we mean consecutive words in the target sequence that are misrecognized; lengths of clusters are measured as the number of tokens, and longer would indicate the more likely one recognition error can cause another. 
From Table \ref{avg_length}, we see a clear distinction between the average length of error clusters for both datasets, that models using only pronunciation for decoder embeddings have them shorter than otherwise. This further shows that, when PET models' decoder embedding only uses pronunciation information, it helps mitigate the problem of error chain reactions.

\begin{table}[]
    \centering
    \caption{Average length of error clusters of recognition results of different Transducer models on AISHELL-2 and THCHS's test datasets. A longer average length would indicate that the model suffers more from error chain reactions.}
    \begin{tabular}{c c c c}
    \toprule
    decoder-emb & joiner-emb & AISHELL & THCHS  \\
    \midrule
    W & W & 1.337 & 1.596 \\
        \midrule
    W & PW &  1.339 & 1.609 \\
    W & PTW &   1.332 & 1.583 \\
    W & CVW & 1.349  & 1.599 \\
    PW & W &   1.319 & 1.614 \\
    \midrule
    P & W & 1.275 & 1.479 \\
    V & W &  1.262 & 1.461 \\
    CV & W & 1.272 & 1.460  \\
    P & PW &  1.265 & 1.468\\
    CV & CVTW & 1.280 & 1.455 \\
         \bottomrule
    \end{tabular}
    \label{avg_length}
\end{table}

\subsection{Recommended PET Configs and Impact on Model Size}

Based on reported experiments and analysis, we would recommend configuring PET models in the following way to optimize results: 1. using a granular representation of pronunciation and no word identity information in the decoder embeddings; 2. use of pronunciation information for the joiner embedding is optional.

Since PET models have multiple embedding tables, it might appear to the readers that, depending on the features used, PET models can potentially have more parameters than the corresponding conventional Transducers. However, before we use PET models for inference, we can pre-compute the final embedding tables rather than storing weights for the individual embedding tables for different features. The final embedding table would have the same size as that of the conventional Transducer, and the same inference speed as well.


\section{Conclusion}

In this paper, we propose Transducers with Pronunciation-aware Embeddings (PET), an extension of the Transducer model that incorporates the pronunciation of text tokens in the model's embedding. 
With experiments on multiple datasets in Mandarin Chinese and Korean, we show that PET can consistently improve ASR accuracy. 

We also discover an interesting phenomenon which we call error chain reactions, where one recognition error is likely to cause other errors to follow.
Our analysis shows that PET can help mitigate the issue of the error chain reaction phenomenon, by reducing the likelihood of producing more errors after a prior error is made.
We believe targeting error chain reaction is a research direction worth more research efforts at a time when the overall ASR accuracy is gradually getting saturated, and we hope this work can cause of chain reaction of more research work in this area.


\section{Acknowledgments}
We thank Taejin Park and Ji Won Yoon for helpful discussions. We also thank Haiyang Sun and Dan Liang.

\bibliography{refs}

\begin{thebibliography}{10}
\providecommand{\url}[1]{#1}
\csname url@samestyle\endcsname
\providecommand{\newblock}{\relax}
\providecommand{\bibinfo}[2]{#2}
\providecommand{\BIBentrySTDinterwordspacing}{\spaceskip=0pt\relax}
\providecommand{\BIBentryALTinterwordstretchfactor}{4}
\providecommand{\BIBentryALTinterwordspacing}{\spaceskip=\fontdimen2\font plus
\BIBentryALTinterwordstretchfactor\fontdimen3\font minus
  \fontdimen4\font\relax}
\providecommand{\BIBforeignlanguage}[2]{{%
\expandafter\ifx\csname l@#1\endcsname\relax
\typeout{** WARNING: IEEEtran.bst: No hyphenation pattern has been}%
\typeout{** loaded for the language `#1'. Using the pattern for}%
\typeout{** the default language instead.}%
\else
\language=\csname l@#1\endcsname
\fi
#2}}
\providecommand{\BIBdecl}{\relax}
\BIBdecl

\bibitem{povey2011kaldi}
D.~Povey, A.~Ghoshal, G.~Boulianne, L.~Burget, O.~Glembek, N.~Goel,
  M.~Hannemann, P.~Motlicek, Y.~Qian, P.~Schwarz, J.~Silovsky, G.~Stemmer, and
  K.~Vesely, ``The {Kaldi} speech recognition toolkit,'' in \emph{ASRU
  Workshop}, 2011.

\bibitem{young2002htk}
S.~Young, G.~Evermann, M.~Gales, T.~Hain, D.~Kershaw, X.~Liu, G.~Moore,
  J.~Odell, D.~Ollason, D.~Povey \emph{et~al.}, ``The htk book,''
  \emph{Cambridge university engineering department}, vol.~3, no. 175, p.~12,
  2002.

\bibitem{graves2006connectionist}
A.~Graves, S.~Fern{\'a}ndez, F.~Gomez, and J.~Schmidhuber, ``Connectionist
  temporal classification: labelling unsegmented sequence data with recurrent
  neural networks,'' in \emph{ICML}, 2006.

\bibitem{graves2012sequence}
A.~Graves, ``Sequence transduction with recurrent neural networks,'' in
  \emph{ICML}, 2012.

\bibitem{chorowski2015attention}
J.~K. Chorowski, D.~Bahdanau, D.~Serdyuk, K.~Cho, and Y.~Bengio,
  ``Attention-based models for speech recognition,'' \emph{NeurIPS}, 2015.

\bibitem{chan2016listen}
W.~Chan, N.~Jaitly, Q.~Le, and O.~Vinyals, ``Listen, attend and spell: A neural
  network for large vocabulary conversational speech recognition,'' in
  \emph{ICASSP}, 2016.

\bibitem{kuchaiev2019nemo}
O.~Kuchaiev, J.~Li, H.~Nguyen, O.~Hrinchuk, R.~Leary, B.~Ginsburg, S.~Kriman,
  S.~Beliaev, V.~Lavrukhin, J.~Cook \emph{et~al.}, ``Nemo: a toolkit for
  building {AI} applications using neural modules,'' \emph{arXiv:1909.09577},
  2019.

\bibitem{ren2019fastspeech}
Y.~Ren, Y.~Ruan, X.~Tan, T.~Qin, S.~Zhao, Z.~Zhao, and T.-Y. Liu, ``Fastspeech:
  Fast, robust and controllable text to speech,'' \emph{NeurIPS}, 2019.

\bibitem{wang2019espresso}
Y.~Wang, T.~Chen, H.~Xu, S.~Ding, H.~Lv, Y.~Shao, N.~Peng, L.~Xie, S.~Watanabe,
  and S.~Khudanpur, ``Espresso: A fast end-to-end neural speech recognition
  toolkit,'' in \emph{ASRU Workshop}, 2019.

\bibitem{ravanelli2021speechbrain}
M.~Ravanelli, T.~Parcollet, P.~Plantinga, A.~Rouhe, S.~Cornell, L.~Lugosch,
  C.~Subakan, N.~Dawalatabad, A.~Heba, J.~Zhong \emph{et~al.}, ``{SpeechBrain}:
  A general-purpose speech toolkit,'' \emph{arXiv:2106.04624}, 2021.

\bibitem{shen2023pronunciation}
P.~Shen, X.~Lu, and H.~Kawai, ``Pronunciation-aware unique character encoding
  for {RNN Transducer-based Mandarin speech recognition},'' in \emph{SLT
  Workshop}, 2023.

\bibitem{9688053}
J.~Yuan, X.~Cai, D.~Gao, R.~Zheng, L.~Huang, and K.~Church, ``Decoupling
  recognition and transcription in mandarin asr,'' in \emph{2021 IEEE Automatic
  Speech Recognition and Understanding Workshop (ASRU)}, 2021, pp. 1019--1025.

\bibitem{wang2021cascade}
X.~Wang, Z.~Yao, X.~Shi, and L.~Xie, ``Cascade rnn-transducer: Syllable based
  streaming on-device mandarin speech recognition with a syllable-to-character
  converter,'' in \emph{2021 IEEE Spoken Language Technology Workshop
  (SLT)}.\hskip 1em plus 0.5em minus 0.4em\relax IEEE, 2021, pp. 15--21.

\bibitem{xu2019improving}
H.~Xu, S.~Ding, and S.~Watanabe, ``Improving end-to-end speech recognition with
  pronunciation-assisted sub-word modeling,'' in \emph{ICASSP}, 2019.

\bibitem{pandey2023procter}
R.~Pandey, R.~Ren, Q.~Luo, J.~Liu, A.~Rastrow, A.~Gandhe, D.~Filimonov,
  G.~Strimel, A.~Stolcke, and I.~Bulyko, ``Procter: Pronunciation-aware
  contextual adapter for personalized speech recognition in neural
  transducers,'' in \emph{ICASSP}, 2023.

\bibitem{chang2021context}
F.-J. Chang, J.~Liu, M.~Radfar, A.~Mouchtaris, M.~Omologo, A.~Rastrow, and
  S.~Kunzmann, ``Context-aware transformer transducer for speech recognition,''
  in \emph{Automatic Speech Recognition and Understanding Workshop (ASRU)},
  2021.

\bibitem{chen2019joint}
Z.~Chen, M.~Jain, Y.~Wang, M.~L. Seltzer, and C.~Fuegen, ``Joint grapheme and
  phoneme embeddings for contextual end-to-end asr.'' in \emph{Interspeech},
  2019.

\bibitem{bruguier2019phoebe}
A.~Bruguier, R.~Prabhavalkar, G.~Pundak, and T.~N. Sainath, ``Phoebe:
  Pronunciation-aware contextualization for end-to-end speech recognition,'' in
  \emph{ICASSP}, 2019.

\bibitem{pundak2018deep}
G.~Pundak, T.~N. Sainath, R.~Prabhavalkar, A.~Kannan, and D.~Zhao, ``Deep
  context: end-to-end contextual speech recognition,'' in \emph{SLT workshop},
  2018.

\bibitem{du2018aishell}
J.~Du, X.~Na, X.~Liu, and H.~Bu, ``Aishell-2: Transforming mandarin {ASR}
  research into industrial scale,'' \emph{arXiv:1808.10583}, 2018.

\bibitem{rekesh2023fast}
D.~Rekesh, S.~Kriman, S.~Majumdar, V.~Noroozi, H.~Juang, O.~Hrinchuk, A.~Kumar,
  and B.~Ginsburg, ``Fast conformer with linearly scalable attention for
  efficient speech recognition,'' \emph{arXiv:2305.05084}, 2023.

\bibitem{wang2015thchs}
D.~Wang and X.~Zhang, ``Thchs-30: A free chinese speech corpus,''
  \emph{arXiv:1512.01882}, 2015.

\bibitem{schmidt2019generalization}
F.~Schmidt, ``Generalization in generation: A closer look at exposure bias,''
  \emph{arXiv:1910.00292}, 2019.

\end{thebibliography}
\end{document}